\documentclass[letterpaper, 10 pt, conference]{ieeeconf}  

\IEEEoverridecommandlockouts                              

\usepackage{url}      
\usepackage{graphicx}
\graphicspath{{03_graphics/}}
\usepackage{amsmath}
\usepackage{amssymb}
\usepackage{algorithm}
\usepackage{algpseudocode}
\usepackage{siunitx}
\usepackage{xcolor,color,soul}
\usepackage{definitions}
\usepackage{macros}
\usepackage[absolute,overlay]{textpos}

\usepackage[colorinlistoftodos, english, disable]{todonotes} 

\usepackage[utf8]{inputenc}

\usepackage[font=footnotesize]{caption}
\usepackage[font=footnotesize]{subcaption}

\usepackage{hyperref}

\title{\LARGE \bf
Decentralized Cooperative Planning for Automated Vehicles with Continuous Monte Carlo Tree Search
}

\author{Karl Kurzer$^{1}$, Florian Engelhorn$^{1}$ and J. Marius Z\"ollner$^{1,2}$ 
	\thanks{
		$^{1}$FZI Research Center for Information Technology, Haid-und-Neu-Str. 10-14, 76131 Karlsruhe, Germany
		{\tt\small zoellner@fzi.de}
		$^{2}$Karlsruhe Institute of Technology, Kaiserstr. 12, 76131 Karlsruhe, Germany
		{\tt\small kurzer@kit.edu, florian.engelhorn@googlemail.com}
	}
}        
        
\begin{document}
\begin{textblock*}{\textwidth}(19mm,10mm)
	\footnotesize
	\noindent\textcopyright2018~IEEE. Personal use of this material is permitted. Permission from IEEE must be obtained for all other uses, in any current or future media, including reprinting/republishing this material for advertising or promotional purposes, creating new collective works, for resale or redistribution to servers or lists, or reuse of any copyrighted component of this work in other works.\\
	\textit{2018 IEEE International Conference on Intelligent Transportation Systems (ITSC)}
\end{textblock*}	
\listoftodos

\maketitle
\thispagestyle{empty}
\pagestyle{empty}

\begin{abstract}
Urban traffic scenarios often require a high degree of cooperation between traffic participants to 
ensure safety and efficiency.
Observing the behavior of others, humans infer whether or not others are cooperating.
This work aims to extend the capabilities of automated vehicles, enabling them to cooperate 
implicitly in heterogeneous environments.
Continuous actions allow for arbitrary trajectories and hence are applicable to a much wider class 
of problems than existing cooperative approaches with discrete action spaces.
Based on cooperative modeling of other agents, Monte Carlo Tree Search (MCTS) in conjunction with 
Decoupled-UCT evaluates the action-values of each agent in a cooperative and decentralized 
way, respecting the interdependence of actions among traffic participants.
The extension to continuous action spaces is addressed by incorporating novel MCTS-specific enhancements for efficient search space exploration.
The proposed algorithm is evaluated under different scenarios, showing that the algorithm is able to achieve effective cooperative planning and generate solutions egocentric planning fails to identify.
\end{abstract}

\section{Introduction}
While the capabilities of automated vehicles are evolving, they still lack an essential
component that distinguishes them from human drivers in their behavior - the ability to
cooperate (implicitly) with others.
Unlike today's automated vehicles, human drivers include the (subtle) actions and intentions of
other drivers in their decisions.
Thus they are able to demand or offer cooperative behavior even without explicit communication.
In recent years many research projects have addressed cooperative driving.
Yet, the focus to date has been on explicit cooperation, which requires communication between 
vehicles or vehicles and infrastructure (\cite{Englund2016, During2014, Frese2007}).

In the foreseeable future neither all vehicles will have the necessary technical equipment to
enable communication between vehicles and the infrastructure, nor will algorithms be standardized to
such an extent that communicated environmental information and behavioral decisions will be
considered uniformly.
Hence, automated vehicles should be able to cooperate with other traffic participants even without 
communication.

\begin{figure}
	\centering
	\def\svgwidth{\columnwidth}
\begingroup%
  \makeatletter%
  \providecommand\color[2][]{%
    \errmessage{(Inkscape) Color is used for the text in Inkscape, but the package 'color.sty' is not loaded}%
    \renewcommand\color[2][]{}%
  }%
  \providecommand\transparent[1]{%
    \errmessage{(Inkscape) Transparency is used (non-zero) for the text in Inkscape, but the package 'transparent.sty' is not loaded}%
    \renewcommand\transparent[1]{}%
  }%
  \providecommand\rotatebox[2]{#2}%
  \newcommand*\fsize{\dimexpr\f@size pt\relax}%
  \newcommand*\lineheight[1]{\fontsize{\fsize}{#1\fsize}\selectfont}%
  \ifx\svgwidth\undefined%
    \setlength{\unitlength}{675.00457764bp}%
    \ifx\svgscale\undefined%
      \relax%
    \else%
      \setlength{\unitlength}{\unitlength * \real{\svgscale}}%
    \fi%
  \else%
    \setlength{\unitlength}{\svgwidth}%
  \fi%
  \global\let\svgwidth\undefined%
  \global\let\svgscale\undefined%
  \makeatother%
  \begin{picture}(1,1.2771669)%
    \lineheight{1}%
    \setlength\tabcolsep{0pt}%
    \put(0,0){\includegraphics[width=\unitlength,page=1]{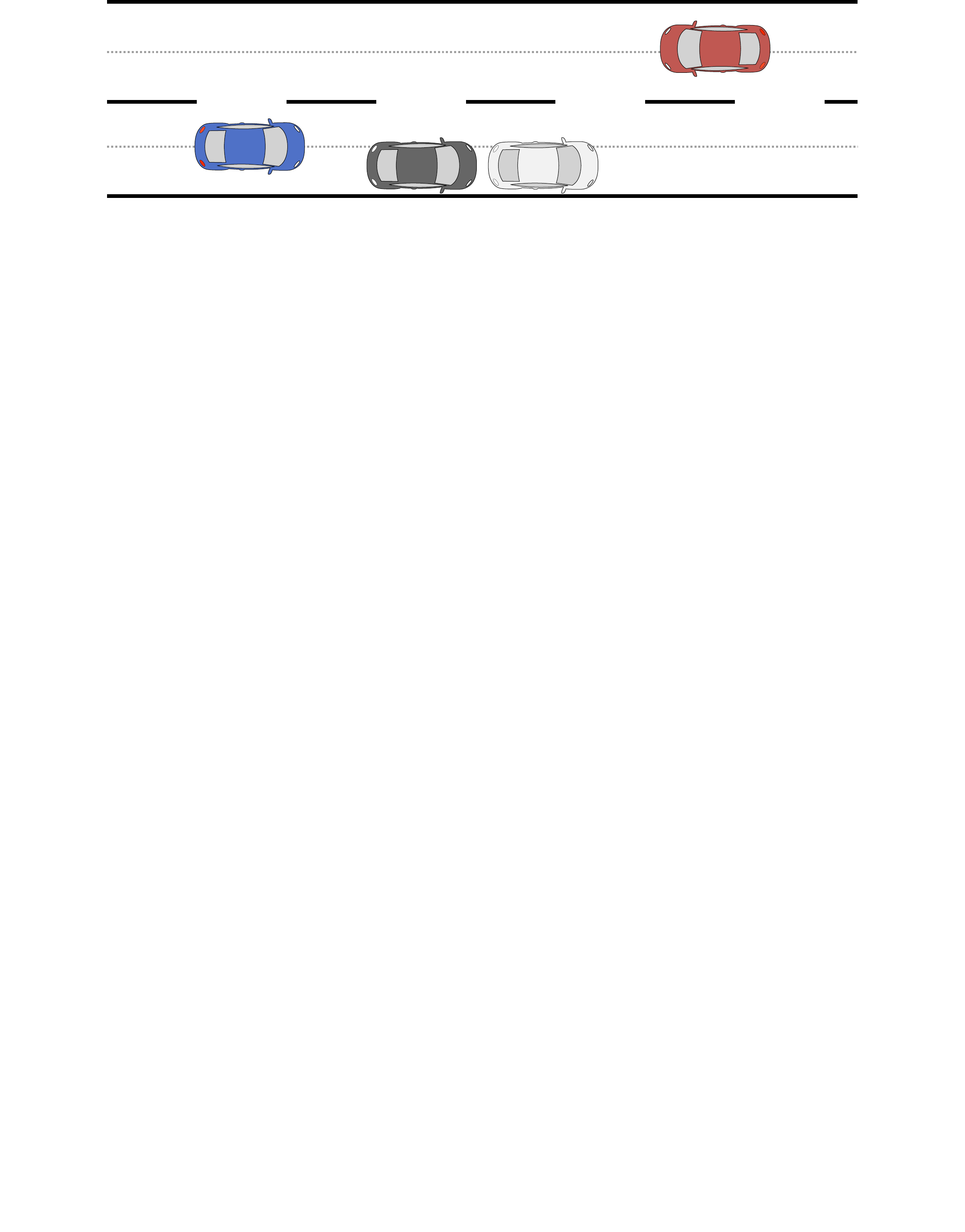}}%
    \put(0.59777307,0.5751626){\color[rgb]{0,0,0}\makebox(0,0)[lt]{\lineheight{0}\smash{\begin{tabular}[t]{l}\fns Backpropagation\end{tabular}}}}%
    \put(0,0){\includegraphics[width=\unitlength,page=2]{main_mcts.pdf}}%
    \put(0.11333269,0.5751626){\color[rgb]{0,0,0}\makebox(0,0)[lt]{\lineheight{0}\smash{\begin{tabular}[t]{l}\fns Simulation\end{tabular}}}}%
    \put(0,0){\includegraphics[width=\unitlength,page=3]{main_mcts.pdf}}%
    \put(0.63999521,1.02221509){\color[rgb]{0,0,0}\makebox(0,0)[lt]{\lineheight{0}\smash{\begin{tabular}[t]{l}\fns Expansion\end{tabular}}}}%
    \put(0,0){\includegraphics[width=\unitlength,page=4]{main_mcts.pdf}}%
    \put(0.12207931,1.02221509){\color[rgb]{0,0,0}\makebox(0,0)[lt]{\lineheight{0}\smash{\begin{tabular}[t]{l}\fns Selection\end{tabular}}}}%
    \put(0,0){\includegraphics[width=\unitlength,page=5]{main_mcts.pdf}}%
  \end{picture}%
\endgroup%

	\caption{Phases of Monte Carlo Tree Search for a passing maneuver;
		the selection phase descends the tree by
		selecting auspicious future states until a state is encountered that has untried actions 
		left.
		After the expansion of the state a simulation of subsequent actions is run until the 
		planning horizon 
		is reached.
		The result is backpropagated to all states along the selected path.
		Ultimately this process converges to the optimal policy.}
	\label{fig:MCTS}
\end{figure}

Automated vehicles nowadays conduct non-cooperative planning, neglecting the interdependence in 
decision-making.
In general this leads to sub-optimal plans and in the worst case to situations that 
can only be mitigated by emergency actions such as braking.
If the perception and decision making by the individual road user is taken into 
account, trajectory sequences can be planned with foresight and safety-critical traffic situations 
can be prevented.

Existing frameworks that integrate prediction and planning to model interdependencies and achieve 
cooperative decision making (\cite{Lawitzky2013, Lenz2016, Bahram2016, Kurzer2018}) are restricted to 
a discrete 
set of actions.
However, the multitude of possible traffic scenarios in urban environments require a holistic 
solution without a priori discretization of the action space.

Thus this work develops a cooperative situation-independent selection of possible actions for each 
traffic participant, addressing situations with a high degree of interaction between road users, 
e.g., road constrictions due to parked vehicles or situations that require to merge into moving 
traffic.

The problem of cooperative trajectory planning is modeled as a multi-agent markov decision process
(MDP) with simultaneous
decision making by all agents \cite{Cowling2012}.
With the help of Monte Carlo Tree Search (MCTS) the optimal policy is inferred.
Monte Carlo Tree Search, a special reinforcement learning method \cite{Vodopivec2017}, has 
shown great potential during multiple occasions facing problems with large branching factors.
AlphaGo, the Go software that reached super-human performance is the most prominent example 
(\cite{Silver2016,Silver2017}).
MCTS improves value estimates of actions by conducting model-based simulations until a terminal state is reached and uses backpropagation to update the nodes along the taken action sequence.
These improved value estimates are essential to direct the selection and expansion phases towards promising areas of the search space.
Browne et al. \cite{Browne2012} present a thorough overview of MCTS and its extensions.
An example for the domain of automated driving is depicted in Fig.~\ref{fig:MCTS}.

The resulting algorithm, \CMCTS\,, plans \textbf{de}centralized \textbf{co}operative trajectories 
sampling \textbf{c}ontinuous 
action spaces.
The problem of decentralized simultaneous decision making is modeled as a matrix game and solved by 
Decoupled-UCT (a variant of MCTS, \cite{Tak2014,Kurzer2018}), removing dependencies on the decisions of others.
In order to cope with the combinatorial explosion resulting from continuous action spaces, we 
enhance the plain MCTS with semantic move groups, kernel updates and guided exploration.
The resulting algorithm is able to perform cooperative maneuvers in tight spaces that are common in 
urban environments. 

This paper is structured as follows:
First, a brief overview of research on cooperative automated driving is given in section II.
Section III introduces the terminology and defines the problem formally.
The general approach to the problem and enhancements to the plain MCTS are presented in Section IV.
Last, \CMCTS's\ extensions for continuous action spaces are evaluated and it's capabilities 
are compared to other planning methods.

\section{Related Work}
Cooperative planning takes the actions, intentions and interdependencies of all traffic 
participants into consideration and seeks to maximize the total utility by following the best 
combined action.
It is important to note that cooperation does not need to be beneficial for each agent (rational 
cooperation), but that it is sufficient if the combined utility increases given a reference point (\cite{During2014, Pascheka2015a}).

One approach that generates cooperative plans for highway scenarios first determines the best 
individual plan using an egoistic driver model.
In case this plan results in a conflict a recursive pairwise conflict resolution process is 
initiated based on the assumption that decisions depend on the traffic ahead.
The algorithm performs an exhaustive search over the available maneuver combinations and the lowest 
non conflicting solution is chosen \cite{Schwarting2014}.

Similarly Düring et al. \cite{During2014} conduct an exhaustive search over a communicated set of 
discrete actions between two vehicles and choose the joint action with the minimum cost.
Building on this, extensions have been developed that aim to incorporate fairness improvements 
avoiding that cooperation becomes single sided \cite{Pascheka2015a}.

Instead of conducting an exhaustive search that quickly becomes intractable for multiple traffic 
participants and longer time horizons, the problem of cooperative decision making has been solved 
employing Monte Carlo Tree Search to estimate the best maneuver combination over multiple time 
steps (\cite{Lenz2016, Kurzer2018}).

Other approaches are not explicitly cooperative, however they do capture the interdependencies of 
actions as they evaluate the threat resulting from different maneuver combinations, and hence 
predict the future motions of vehicles \cite{Lawitzky2013} and are able to generate proactive 
cooperative driving actions \cite{Bahram2016}.

Furthermore, off-line calculated maneuver templates can be used to devise cooperative plans. 
First, the maneuver template needs to match a given traffic scene with specific initial 
constraints, then the maneuver described in the template is checked for feasibility.
If multiple templates are feasible given a specific traffic scene the maneuver template with the 
lowest cost specifies the cooperative maneuver \cite{Manzinger2017}.

While the aforementioned approaches do conduct cooperative planning and most of the 
approaches consider the interdependency between the decisions of the individual traffic 
participants, some require communication and none are able to plan cooperative maneuvers for 
arbitrary action spaces that are required by obstructed road layouts and tight urban spaces.

\section{Problem Statement}
The problem of cooperative trajectory planning is formulated as a decentralized Markov Decision 
Process (Dec-MDP).
Agents independently choose an action in each time step without knowing the decisions taken
by others. 
Each agent collects an immediate reward and the system is transfered to the next state.
Being a cooperative multi-agent system the state transition as well as the reward are dependent on 
all agents' actions.

The Dec-MDP is described by the tuple $\langle \agentspace, \statespace,  \actionspace, \transitionmodel, \reward, \discountfactor\rangle$, presented in \cite{Kurzer2018}.

\begin{itemize}
	\item $\agentspace$ denotes the set of \emph{agents} indexed by $i \in {1, 2, \dots n}$. 
	\item $\statespacei$ denotes the \emph{state space} of an agent, $\statespace = \times 
	\statespacei$ represents the joint state space of $\agentspace$. 
	\item $\actionspacei$ denotes the \emph{action space} of an agent, $\actionspace= \times 
	\actionspacei$ represents the joint action space of $\agentspace$. 
	\item $\transitionmodel: \statespace \times \actionspace \times \statespace \to [0,1]$ is the
	\emph{transition function} $P(s' | s, \actionj)$ specifying the probability of a 
	transition from state $s$ to state $s'$	given the joint action $\actionj$ chosen by each 
	agent independently.
	\item $\reward: \statespace \times \actionspace \times \statespace \to \mathbb{R}$ is the 
	reward function with $r(s, s',\actionj)$ denoting the resulting reward of the joint action 
	$\actionj$.
	\item $\discountfactor \in [0,1]$ denotes a \emph{discount factor} controlling the influence of 
	future rewards on the current state. 
\end{itemize}

The superscript $^i$ is used to indicate that a parameter relates to a specific agent $i$.
The joint policy $\Pi = \langle \pi^1, \dots, \pi^n \rangle$ is a solution to the cooperative 
decision problem. An individual policy, $\pi^i$, for a single agent, i.e., a mapping from the state 
to the probability of each available action is given by $\pi^i: \statespace^i\times 
\actionspace^i \to [0,1]$. 

It is the aim of each agent to maximize its expected cumulative reward in the MDP, starting from its current state:
$G = \sum {\gamma^t r(s ,s', a)}$ where $t$ denotes the time and $G$ the return, representing the 
cumulated discounted reward.
$V(s)$ is called the state-value function, given by $V^\pi(s) = E[G|s,\pi]$.
Similarly, the action-value function $Q(s,a)$ is defined as $Q^\pi(s,a) = E[G|s,a]$,
representing the expected return of choosing action $\action$ in state $\state$.

The optimal policy starting from state $s$ is defined as $\pi^* = \argmax_{\pi} V^\pi(s)$. 
The state-value function is optimal under the optimal policy: $max\ V = V^{\pi^*}$, the same is 
true for the action-value function: $max \ Q=Q^{\pi^*}$.
The optimal policy is found by maximizing over $Q^*(s,a)$:
\begin{equation}\label{Eq:OptPolicy}
\pi^{*}(a|s) = \left\{ \begin{array}{rcl}
1 & {\mbox{if}} \ {a = \argmax_{a \in {\cal A}} Q^{*}(s,a)}  \\
0 & \mbox{otherwise}
\end{array}\right.
\end{equation}
The optimal policies can easily be derived, once $Q^*$ has been determined.
Hence the goal is to learn the optimal action-value function $Q^*(s,a)$ for a given state-action combination.

\section{Approach}
While discrete actions as opposed to classical trajectory planning allows to plan over longer
periods of time \cite{Kurzer2018}, the resolution is insufficient to plan detailed maneuvers.
Due to the multitude of possible traffic scenarios in urban environments, a solution with heuristic 
a priori discretization of the action space of road users is not suitable (\cite{During2014, Lenz2016, Bahram2016, Kurzer2018}).
The planning of safe maneuvers within such a dynamic environment can only take place if the
trajectory planning is equipped with a situation-independent selection of possible actions for each
road user.
In order to address driving situations with a high degree of interaction between the road users 
involved such as road constrictions due to parked vehicles, we employ a MCTS-based approach similar 
to (\cite{Kurzer2018,Lenz2016}) but extend it to continuous action spaces.
By extending the method, any desired trajectory sequence can be generated in order to solve complex 
scenarios through cooperative behavior.

The following subsections first explain the action space and the validation of actions. Based on 
actions and the resulting state the cooperative reward is described in the following subsection. 
The last subsections shine some light on the enhancements for continuous action spaces that 
enable a structured exploration and thus quicker convergence.

\subsection{Action Space}
Actions are applied to the vehicle's state which is given by its position, velocity and acceleration in
longitudinal and lateral direction as well as its heading.
An action is defined as a pair of values with a longitudinal velocity change $\Delta 
v_{longitudinal}$ and a lateral position change $\Delta y_{lateral}$.
The value pair describes the desired change of state to be achieved during the action duration 
$\Delta T = t_1 - t_0$.
Using this pair of values in combination with the following initial and terminal conditions for 
the longitudinal as well as lateral direction, quintic polynomials are solved to generate 
jerk-minimizing trajectories for the respective direction \cite{Takahashi1989}.

The initial conditions for the longitudinal as well as lateral position, velocity and acceleration are determined by the current state. The terminal conditions are defined with:
\begin{align}
&\ddot{x}(t_1) = 0\\
&\dot{y}(t_1) = 0\\
&\ddot{y}(t_1) = 0
\end{align}

The initial and terminal conditions leave three free constraints with $\dot{x}(t_1)$,  $y(t_1)$ and $x(t_1)$. 
The desired velocity change in longitudinal direction $\Delta v_{longitudinal}$ \eqref{Eq:LateralPosition}, as well as the lateral offset of the trajectory $\Delta y_{lateral}$ \eqref{Eq:LongitudinalVelocity} are parameterized and introduced as dimensions 
of the action space of each agent.
\begin{align}\label{Eq:LongitudinalVelocity}
	\dot{x}(t_1) &= \dot{x}(t_0) + \Delta v_{longitudinal}
\end{align}

\begin{align}\label{Eq:LateralPosition}
	y(t_1) = y(t_0) + \Delta y_{lateral}
\end{align}

The last free unknown is the distance covered in longitudinal direction, \eqref{Eq:LongitudinalPosition}.
In order to describe a physically feasible maneuver, it is defined as
\begin{equation}\label{Eq:LongitudinalPosition}
x(t_1) = \frac{\dot{x}(t_0) + \dot{x}(t_1)}{2} \Delta T
\end{equation}

\subsection{Action Validation}
The resulting trajectory needs to be drivable for a front axle-controlled vehicle and should be 
directly trackable by a trajectory controller.
Hence, we need to conduct an action validation, using kinematic as well as physical 
boundary conditions so that the simulation adheres to the constraints.

The continuity of the curvature, the steering angle as well as the vehicle dependent permissible minimal radii ensure that the resulting 
trajectories are drivable.
In addition, the dynamic limits of power transmission limit the maximum and minimum acceleration of 
the vehicle.

\subsection{Cooperative Reward Calculation}
The immediate individual reward based on the behavior of an agent is calculated using 
\eqref{Eq:Reward}.
It considers the states reached and the actions taken, as well as a validation reward.
\begin{equation}\label{Eq:Reward}
r^i = r^i_{state} + r^i_{action} + r^i_{validation}
\end{equation}
The state reward $r^i_{state}$ is determined given the divergence between the current and the 
desired state.
A desired state is characterized by a longitudinal velocity $v_{des}$ and a lane index $k_{des}$.
To ensure that the agent drives in the middle of a lane, deviations from the lane's center line 
$\Delta y$ inflict penalties.

Actions always result in negative rewards (costs).
They create a balance between the goal of minimizing the deviation from the desired state as 
quickly as possible and the most economical way to achieve this.
Currently, $r^i_{action}$ considers only basic properties such as the longitudinal and lateral 
acceleration \eqref{Eq:LongitudinalAcceleration} and \eqref{Eq:LateralAcceleration} as well as lane 
changes, \eqref{Eq:LaneChange}.
However, they can easily be extended to capture additional safety, efficiency and comfort related 
aspects of the generated trajectories.
The order of importance is adjusted with the respective weights.
\begin{equation}\label{Eq:LongitudinalAcceleration}
C_{\ddot{x}} = w_{\ddot{x}} \int_{t_0}^{t_1} (\ddot{x}(t))^2 dt
\end{equation}

\begin{equation}\label{Eq:LateralAcceleration}
C_{\ddot{y}} = w_{\ddot{y}} \int_{t_0}^{t_1} (\ddot{y}(t))^2 dt
\end{equation}

\begin{equation}\label{Eq:LaneChange}
C_{k\pm} = w_{k\pm}
\end{equation}

The last term is the action validation reward, see \eqref{Eq:ActionValidation}.
It evaluates whether a state and action is valid, i.e., being inside the drivable environment and 
adhering to the the kinematic as well as physical constraints and whether a state 
action combination is collision free.
\begin{equation}\label{Eq:ActionValidation}
r^i_{validation} = r^i_{invalid\,state} + r^i_{invalid\,action} + r^i_{collision}
\end{equation}

To achieve cooperative behavior a cooperative reward \cooperativeReward\ is defined.
The cooperative reward of an agent $i$ is the sum of its own rewards, see \eqref{Eq:Reward}, as 
well as the sum of all other rewards of all other agents multiplied by a cooperation factor 
$\lambda^i$, see \eqref{Eq:CooperativeReward}, (\cite{Lenz2016,Kurzer2018}).
The cooperation factor determines the agent's willingness to cooperate with other agents (from 
$\lambda^i = 0$ \emph{egoistic}, to $\lambda^i = 1$ \emph{fully cooperative}).
\begin{equation}\label{Eq:CooperativeReward}
\cooperativeReward = r^i + \lambda^i \sum_{j = 0, j \neq i}^{n}r^{j}
\end{equation}

\subsection{Progressive Widening}
In the basic version of the Monte Carlo Tree Search, agents are modeled with discrete action 
spaces of constant size.
Since each action must be visited at least once when using the UCT algorithm \cite{Kocsis2006}, the 
method cannot simply be applied to continuous action spaces. 
Using \textit{Progressive Unpruning} \cite{Chaslot2008} or \textit{Progressive Widening} 
\cite{Coulom2007} the number of discrete actions within the action space can be gradually 
increased at runtime. 
A larger number of available actions increases the branching factor of the search tree.

At the beginning, the agent receives an initial, discrete set of available actions in each state. 
If a state has been visited sufficiently often, it gets progressively widened, by adding another 
action to the action space.
Note that progressive widening is conducted on a per agent basis.
The criterion for progressive widening is defined as: 
\begin{equation} \label{eq: progressive widening criteria}
N(A(s)) \geq C_{PW} \cdot n(s)^{\alpha_{PW}}
\end{equation}
The number of possible actions $N(A(s))$ in state $s$ therefore depends directly on the visit count 
of the state $n(s)$. 
The parameters $C_{PW}$ and $\alpha_{PW} \in[0,1]$ must be adapted empirically to the respective 
application.

The simplest way to add new actions is to select a random action from the continuous action space. 
More advanced approaches use the information available in the current state, so that promising 
areas within the action space can be identified and new actions can be added 
(\cite{Couetoux2012, Yee2016}).

\subsection{Semantic Action Grouping}
The following section introduces the concept of semantic action grouping. 
The approach taken is similar to the concepts of (\cite{Saito2007, Childs2008}).

The goal of grouping similar actions is to reduce the computational complexity and increase the robustness of the algorithm. 
The grouping of actions is connected with the implementation of heuristic application knowledge. 

Each action is uniquely assigned to a state-dependent action group using a criterion. 
During the update phase, the characteristic values of the groups are calculated on the basis of the 
characteristic values of the group members.
In the selection strategy, the best action group according to UCT is first selected using the group 
statistics before a specific action is identified within the group. 
The action groups serve as filters for determining relevant areas of the action space. 
Within the action groups, the information of the individual actions is generalized to the 
group-specific values, but remains unaffected. 
By using an unambiguous assignment function from the action space to the group action space it is 
ensured that the action groups do not overlap.

\begin{figure}
	\centering
	\def\svgwidth{0.8\columnwidth}
	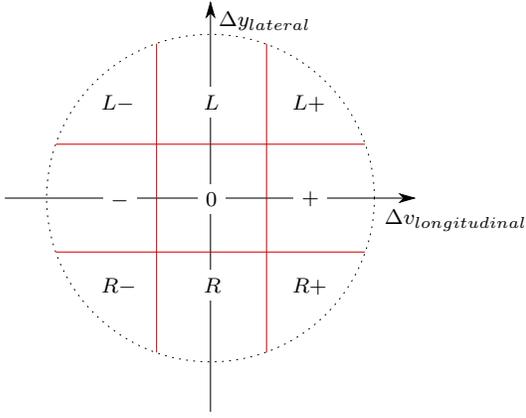
	\caption{Segmentation of the action space into semantic action groups; The semantic action 
		groups are dependent on the successor state of the agent's current state, with $L$ and $R$ 
		denoting lane changes to the left and right lane and $+$ and $-$ acceleration and 
		deceleration respectively.}
	\label{fig:ActionClasses}
\end{figure}

Semantic action groups describe state-dependent, discrete areas within an agent's action space. 
The areas divide the action space into the following nine areas based on the semantic state 
description of the agent's future state, as depicted in Fig.~\ref{fig:ActionClasses}:\\

\begin{itemize}
	\setlength\itemsep{-0.5em} 
	\item \textit{No/slight change of state: $0$} \\

	\item \textit{accelerate: $+$} \\
	
	\item \textit{decelerate: $-$} \\
	
	\item \textit{decelerated lane change to the left: $L-$} \\
	
	\item \textit{lane change to the left: $L$} \\
	
	\item \textit{accelerated lane change to the left : $L+$} \\
	
	\item \textit{decelerated lane change to the right: $R-$} \\
	
	\item \textit{lane change to the right: $R$} \\
	
	\item \textit{accelerated lane change to the right: $R+$} \\
	
\end{itemize}

\subsection{Similarity Update}
\begin{figure}
	\centering
	\def\svgwidth{0.8\columnwidth}
	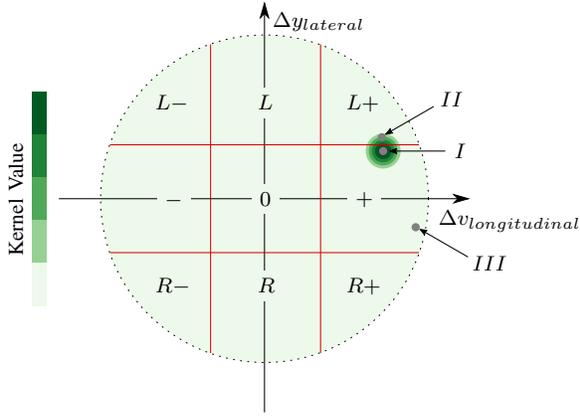
	\caption{Similarity update between two actions from different semantic action groups; In the 
	update phase the result from action $I$ is used to update actions that are within within the 
	kernel, i.e. $II$, and thus lead to a similar subsequent state.}
	\label{fig:SimilarityUpdate}
\end{figure}

As shown in Fig.~\ref{fig:SimilarityUpdate}, actions can generate similar ego states 
independently of the semantic description of their subsequent state and the resulting action group. 
For example, the difference between the next state of action $I$ and $II$ is a different track 
assignment.
Although this describes different semantic states, the lateral position deviates only slightly. 
In contrast, action $III$ is assigned to the same action group as action $I$, but has a much larger 
difference with regard to the lateral deviation. 
To overcome the restrictions of non-overlapping semantic groups, the values of an agent's 
actions are generalized to similar actions in the update step, regardless of the action 
groupings and the resulting boundaries within the action space. 
During the backpropagation phase of MCTS, the similarity between the current action $a'$ and all previously explored actions of the agent at the expanded state is determined using a distance measure. 
This measure of similarity is then used to weight the result of the current sample and select similar states to update. 
The similarity measure is calculated as


\begin{equation}
K(a,a') = \exp \left ( -\gamma \left \Vert a - a' \right \Vert^2 \right ) \forall a \in A_{exp}(s)
\end{equation}
With the help of the radial basis function a symmetrical distance dimension in the value range 
$K(a_i,a^*) \in[0,1]$ can be mapped. 
The different dimensions of the action space can also be weighted differently. 
Since both dimensions are of the same order of magnitude, the same weighting is used for both dimensions.
Due to the continuous nature of the kernel the visit count of states is no longer an integer,
but a floating point number.

\subsection{Guided Search}
In \cite{Couetoux2012} a heuristic -- the so-called \textit{Blind Values} (BV) -- $n$ randomly drawn actions of the theoretical action space $A$ are evaluated and the action with the resulting maximum Blind Value, see \eqref{Eq:BlindValueMax},
is added to the locally available action space of the state.
Blind values make use of the information gathered on all actions already explored from a given state in the past to select promising areas of exploration.
\begin{equation}
    \label{Eq:BlindValueMax}
	a^* =  \argmax_{a' \in A_{rnd}} \text{BV}(a',\rho,A_{exp}) 
\end{equation}
The basic idea is to select actions at the beginning, which have a large distance to previous 
actions in order to foster exploration, avoiding local optima. 
However, as the visit count increases, actions with a short distance to actions with high 
UCT values are preferred.
To regularize the sum of both criteria, the statistical properties of the previous data points -- 
mean value and standard deviation -- are used.

The blind value as a measure of the attractiveness of an action is calculated with
\begin{align} \label{eq: blind value}
\text{BV}(a',\rho,A_{exp}) &= \underset{a\in A_{exp}}{\min} \text{UCT}(a) + \rho \cdot K(a,a')
\end{align}	
where
\begin{align}
\rho &= \frac{\sigma_{a\in A_{exp}}(\text{UCT}(a))}{\sigma_{a'\in A_{rnd}}(K(0,a'))}
\end{align} 
\begin{align*}
0&: \text{center of action space}\\
K(a,a')&: \text{distance function for two actions}\\
A_{exp}&: \text{set of discrete, previously explored actions}\\
A_{rnd}&: \text{set of discrete, randomly selected actions}
\end{align*}
So that the current statistics of the agent and its action space are being considered.


\section{Evaluation}
\begin{figure*}
	\begin{subfigure}{\columnwidth}
		\centering
		\includegraphics[width=0.5\textwidth]{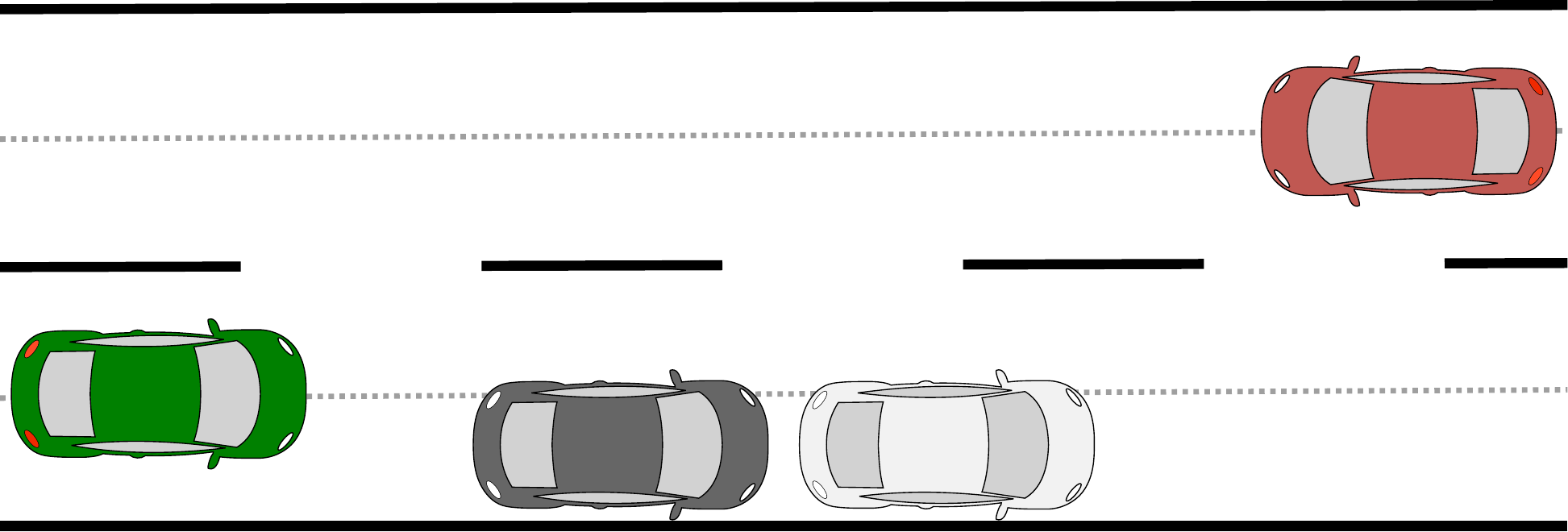}
		\caption{Bottleneck scenario; gray vehicles are parked, the red and green vehicles can only 
		pass at the same time if the red vehicle cooperates by moving to the right}
		\label{fig:Bottleneck}
	\end{subfigure}
	\hspace{0.5cm}
	\begin{subfigure}{\columnwidth}
		\centering
		\includegraphics[width=0.5\textwidth]{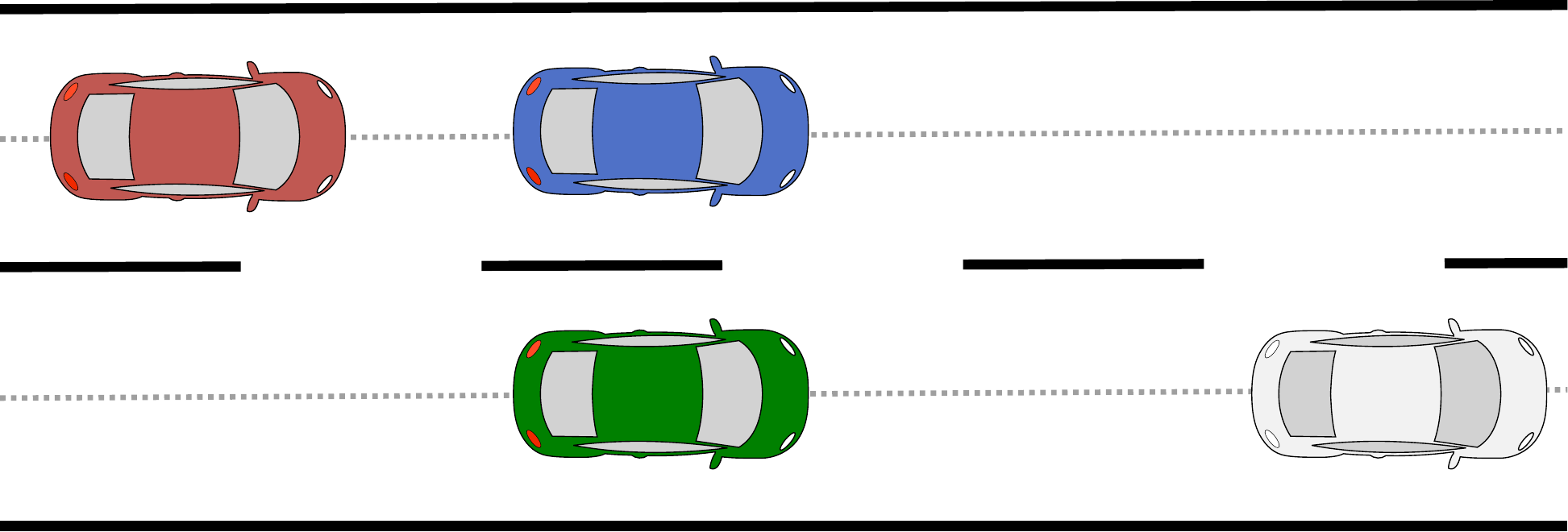}
		\caption{Merge-in scenario; the gray vehicle is parked, the green vehicle can either, 
		accelerate, passing first, merge-in between the red and the blue vehicle, or decelerate to 
		pass last}
		\label{fig:Merge-in}
	\end{subfigure}
	\caption{Two evaluation scenarios}
	\label{fig:Scenarios}
\end{figure*}

The evaluation is conducted using a simulation.
The developed algorithm is evaluated under two different scenarios, namely \emph{bottleneck} and 
\emph{merge-in}, (see Fig.~\ref{fig:Scenarios}).
First, the enhancements with regard to continuous actions are evaluated.
Second, we demonstrate that our algorithm can achieve effective cooperative planning and generate 
solutions egocentric planning fails to identify.
A video of the algorithm in execution can be found online 
\footnote{http://url.fzi.de/DeCoC-MCTS-ITSC}.

\subsection{Search Space Exploration}
To get a better understanding of the effects of the enhancements and their combinations, the exploration of the agent's action space is evaluated for the root node. 
This is done using the bottleneck scenario, depicted in Fig.~\ref{fig:Bottleneck}.

\begin{figure}
	\begin{subfigure}{\columnwidth}
		\centering
		\includegraphics[clip, trim= 0cm 0cm 1.8cm 		
		0cm,width=0.68\textwidth]{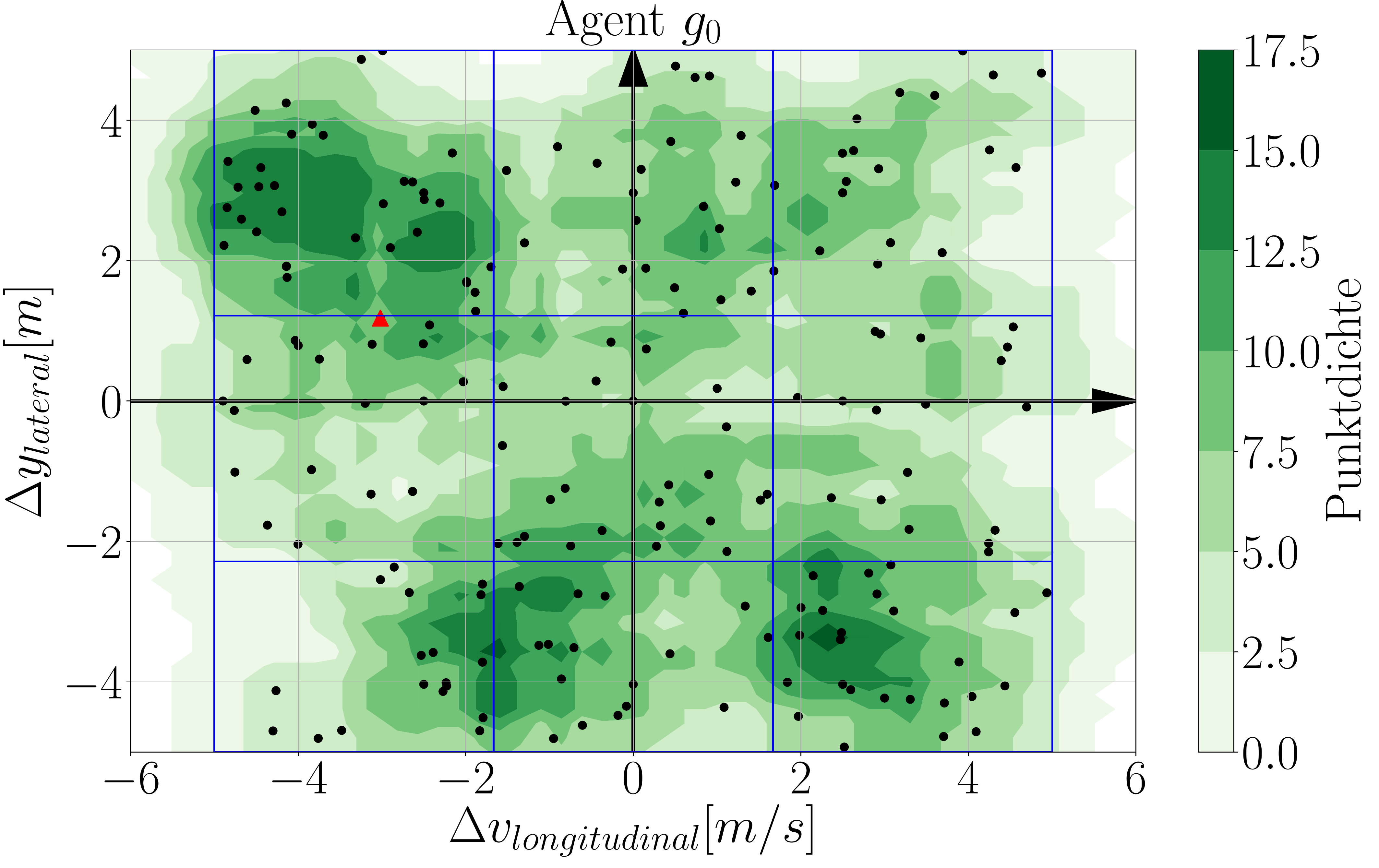}
		\caption{Basic algorithm}
		\label{fig:Basic}
	\end{subfigure}
	\begin{subfigure}{\columnwidth}
		\centering
		\includegraphics[clip, trim= 0cm 0cm 1.8cm 		
		0cm,width=0.68\textwidth]{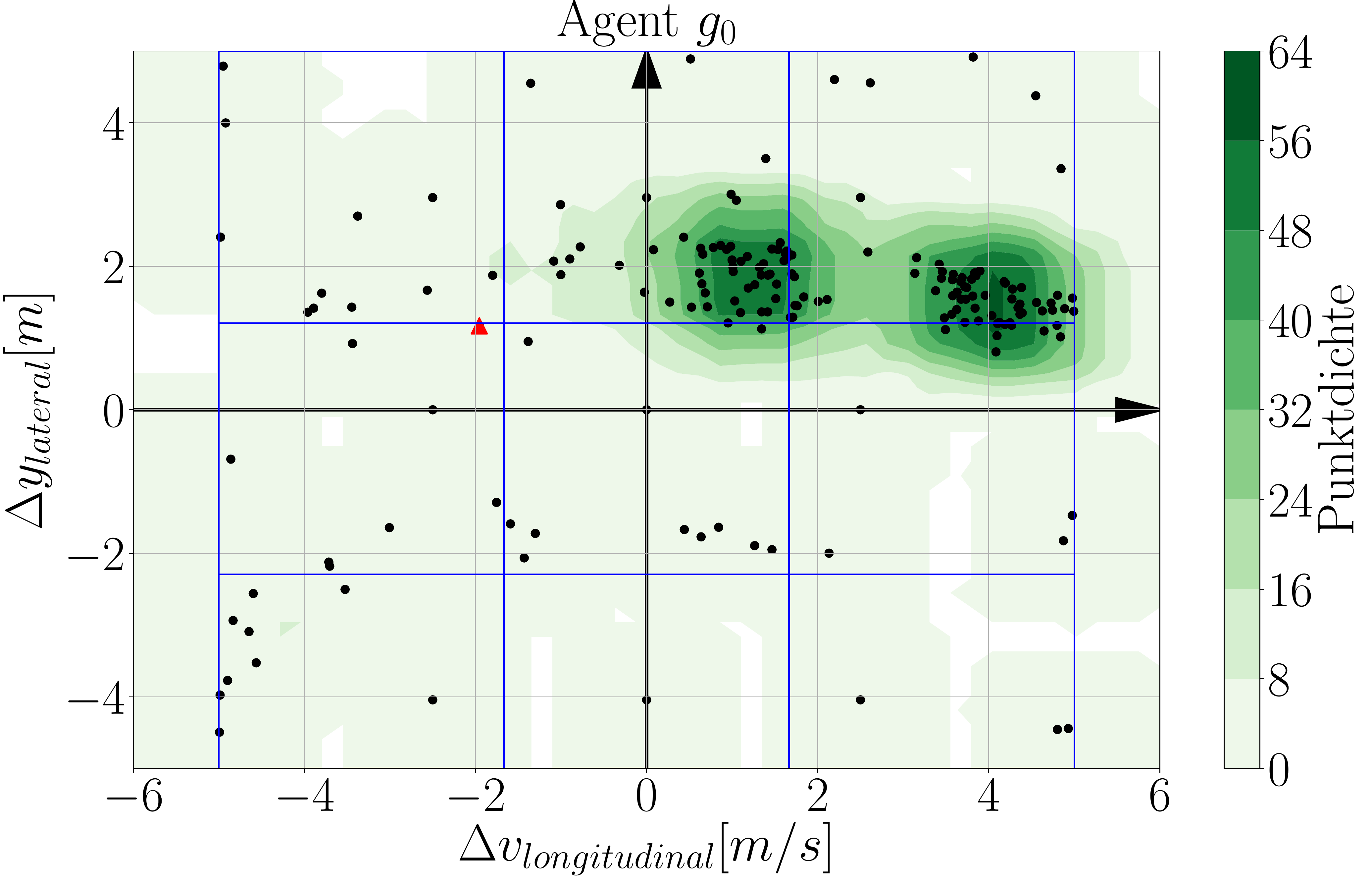}
		\caption{Basic with guided search}
		\label{fig:Guided}
	\end{subfigure}
	\begin{subfigure}{\columnwidth}
		\centering
		\includegraphics[clip, trim= 0cm 0cm 1.8cm 		
		0cm,width=0.68\textwidth]{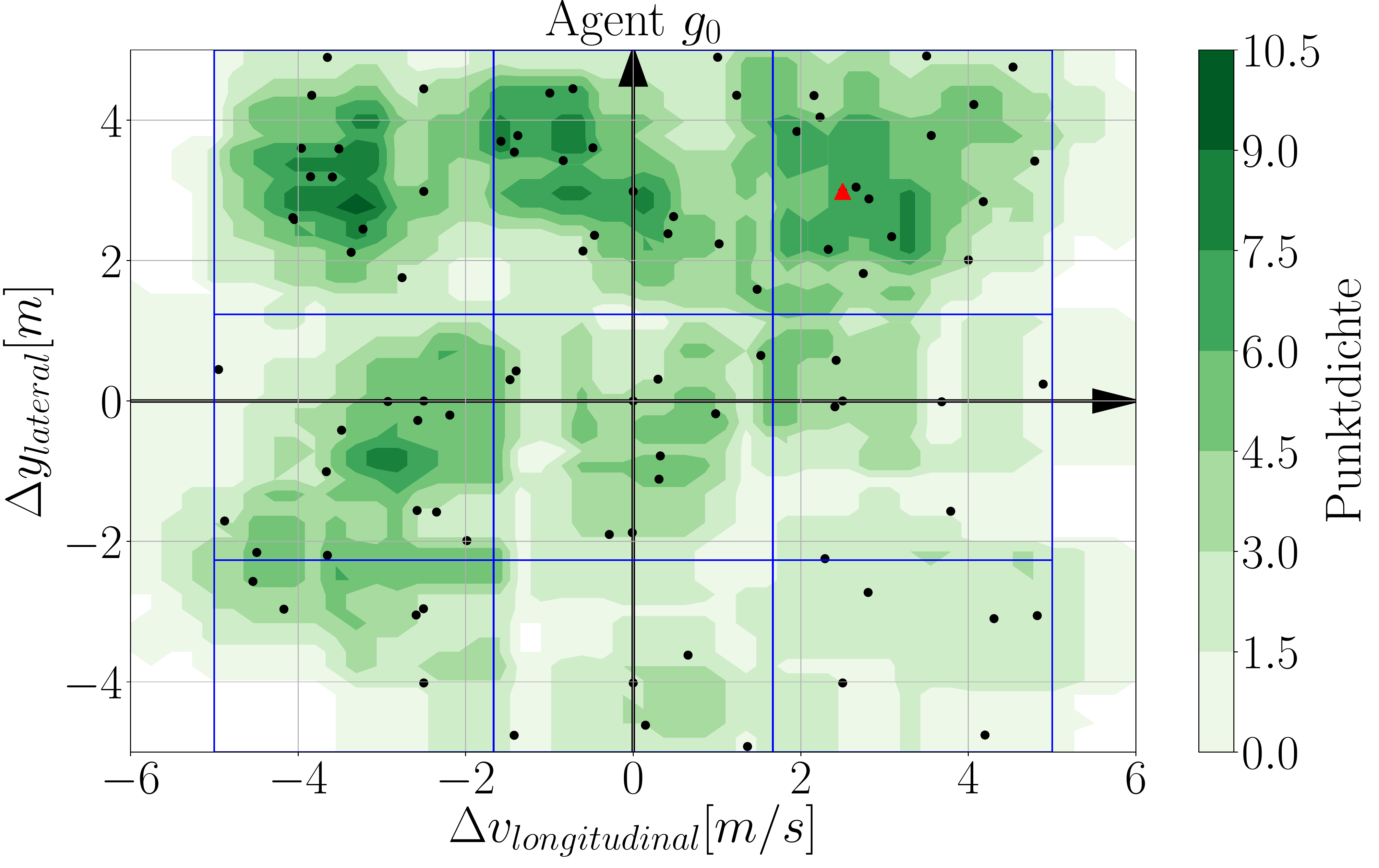}
		\caption{Semantic action grouping}
		\label{fig:MoveGroups}
	\end{subfigure}
	\begin{subfigure}{\columnwidth}
		\centering
		\includegraphics[clip, trim= 0cm 0cm 1.8cm 		
		0cm,width=0.68\textwidth]{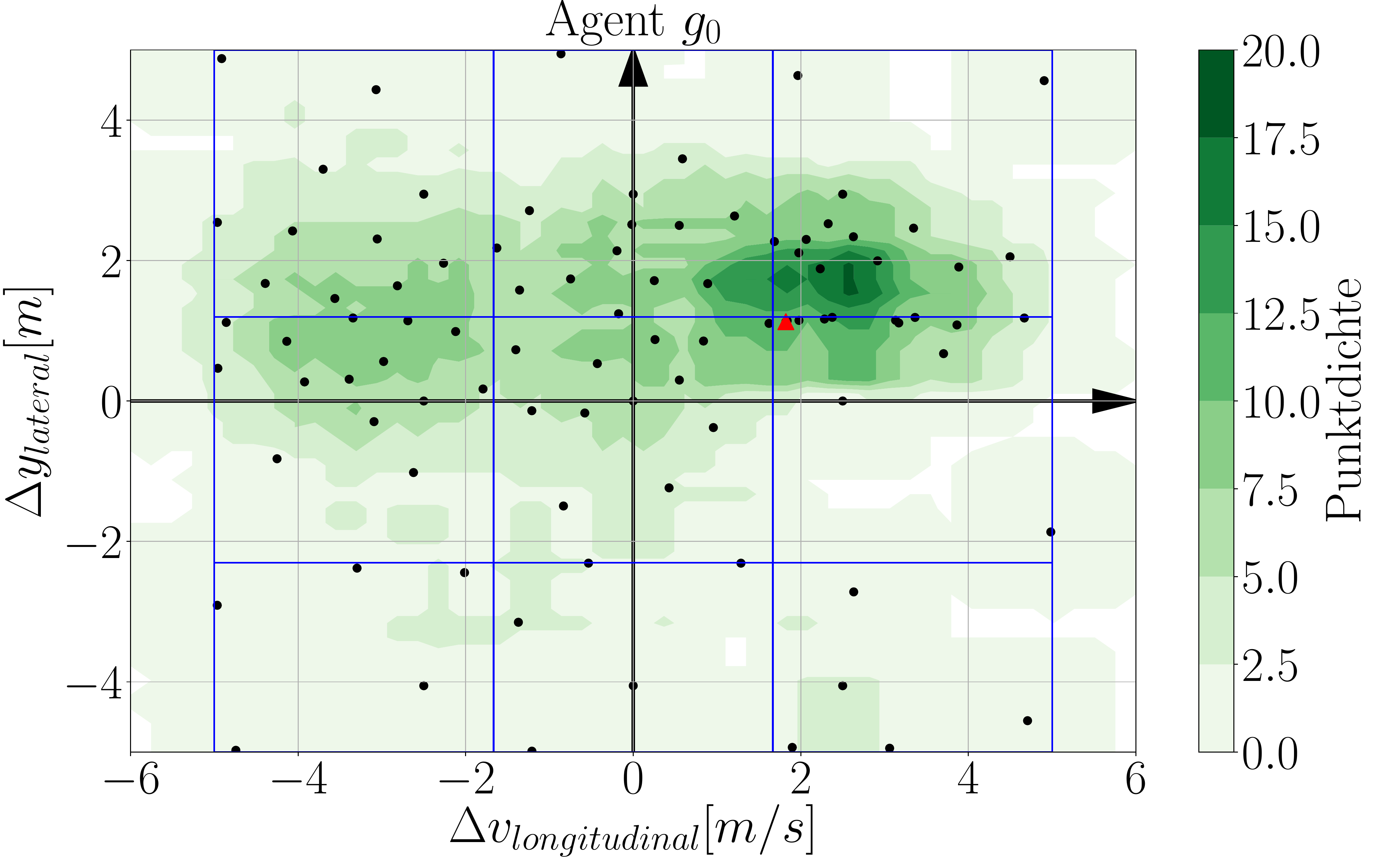}
		\caption{Semantic action grouping with guided search}
		\label{fig:GuidedMoveGroups}
	\end{subfigure}
	\begin{subfigure}{\columnwidth}
		\centering
		\includegraphics[clip, trim= 0cm 0cm 1.8cm 		
		0cm,width=0.68\textwidth]{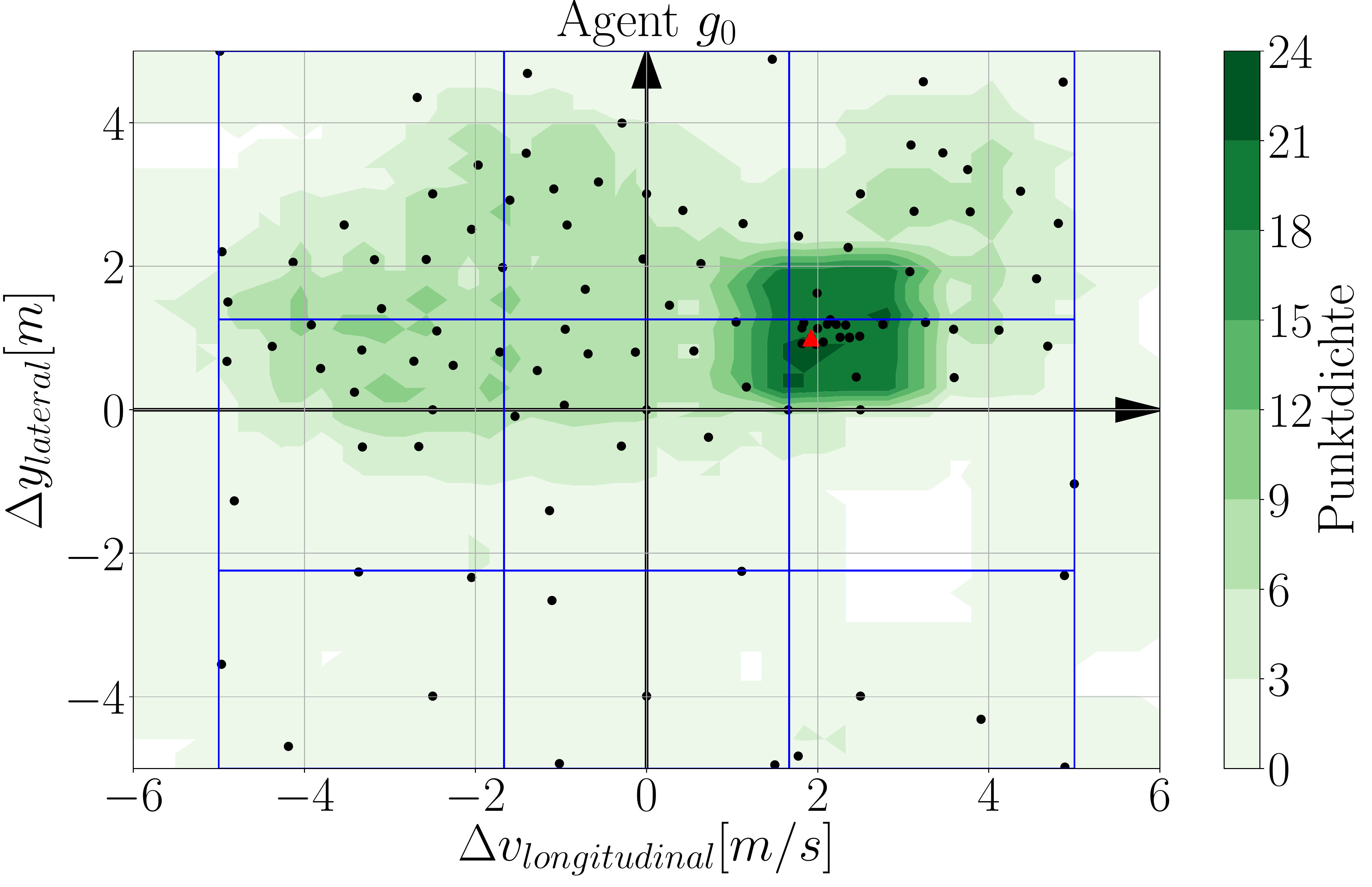}
		\caption{Semantic action grouping with guided search and similarity update}
		\label{fig:GuidedMoveGroupsSimilarity}
	\end{subfigure}
	\caption{Comparison of the search space exploration of the green agent during a lane change in 
	the bottleneck scenario, Fig.~\ref{fig:Bottleneck}
	Each sample marks an explored action.
	The color scale shows the sample density.
	The state limits introduced by the semantic action groups are shown in blue.
	The red triangle marks the selected action.
	The selection is made on the basis of the highest action-value.}
	\label{fig:ExplorationAnalysis}
\end{figure}

Shortly after the start of the scenario, the green agent changes to the left for all variants of 
the enhancements.
This point is used to analyze the exploration of the action space exemplarily.
The actions available to the green agent are shown in Fig.~\ref{fig:ExplorationAnalysis}.
The action with the highest action-value is selected for execution (red triangle).

The basic MCTS algorithm explores the search space randomly, (Fig.~\ref{fig:Basic}).
There is no connection between the distribution of samples and the selected action.
In comparison, the combination with the guided search results in extreme accumulations of actions 
in the area of an accelerated lane change, (Fig.~\ref{fig:Guided}). 
Other areas of the action space are neglected.
The selected action lies in an unexplored area of the action space.

Using semantic action groups the result clearly differs, (cf. Fig.~\ref{fig:MoveGroups}),
as the exploration is more structured.
The number of samples in the action space is heavily reduced.
Within each semantic action group, actions are randomly distributed.
The action groups for lane changes to the left are visited more frequently and
one of these action groups entails the final selection.

If semantic action groups are additionally combined with the guided search, 
(see Fig.~\ref{fig:GuidedMoveGroups}), samples accumulate in the area of a lane change 
to the left.
The selected action is close to the highest action density.

Adding the similarity update to the previous configuration a similar behavior can be observed, 
(Fig.~\ref{fig:GuidedMoveGroupsSimilarity}).
In addition, there is a strong focus on the target region in form of a high density of possible 
actions.
The selected action is contained in this densely explored area.

The combination of all extensions achieves a structured exploration, requiring less samples due to its higher efficiency than the basic algorithm, indicating the potential of the methods introduced.

\subsection{Scenarios}
Using the scenarios depicted in Fig.~\ref{fig:Scenarios} \CMCTS\ is evaluated for two different 
predictions.
Namely, that other vehicles do not cooperate and keep their velocity (constant velocity assumption) as well as that other vehicles will cooperate.
We conduct a qualitative evaluation based on the velocity deviation of all vehicles in a scene,
where lower overall deviations indicate more efficient driving and a higher total utility.

\begin{figure}
	\centering
	\includegraphics[clip, trim= 1.2cm 1cm 2.9cm 		
	1cm,width=0.95\columnwidth]{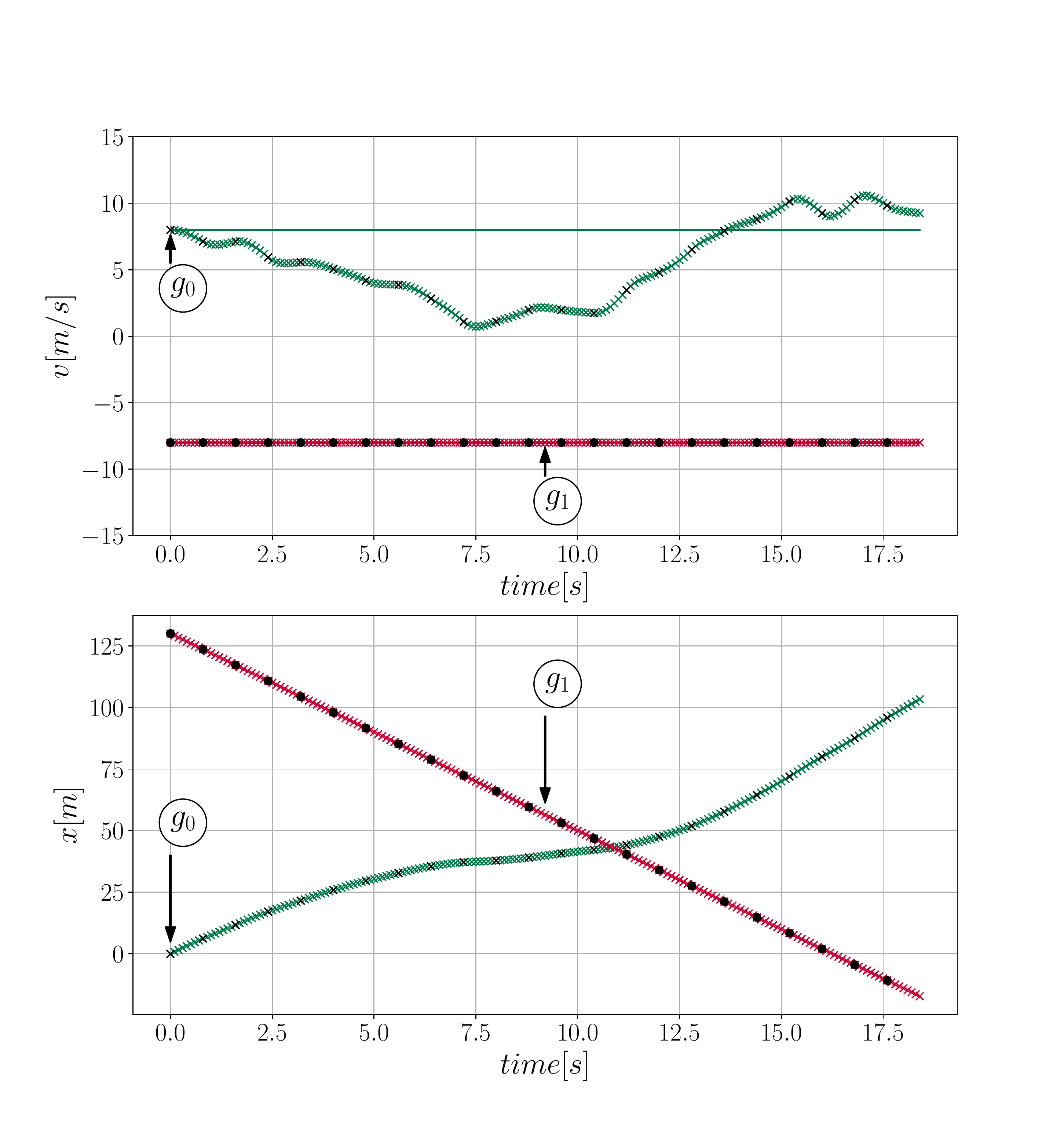}
	\caption{Bottleneck scenario with constant velocity assumption;
		agent $g_0$ reduces its velocity heavily, while agent $g_1$ approaches the obstacle with 
		its desired velocity.}
	\label{fig:bottleneck_uncooperative}
\end{figure}

\begin{figure}
	\centering
	\includegraphics[clip, trim= 1.4cm 1cm 2.9cm 		
	1cm,width=0.95\columnwidth]{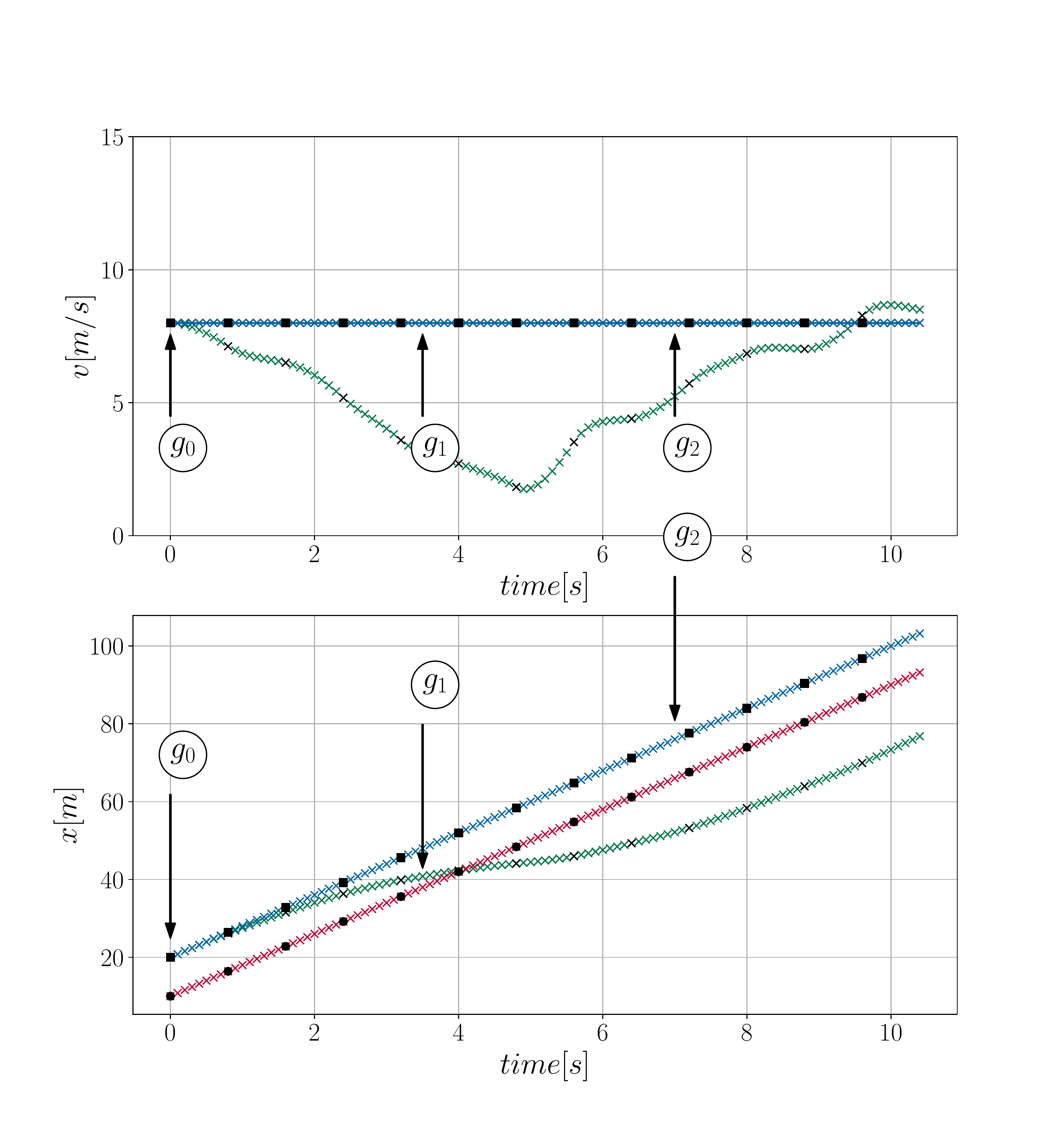}
	\caption{Merge-in scenario with constant velocity assumption;
		agent $g_0$ reduces its velocity heavily, while agent $g_1$ and $g_2$ continue with their 
		desired velocity}
	\label{fig:merge_in_uncooperative}
\end{figure}

The first is the defensive constant velocity assumption that is common in the vast amount of 
prediction and planning frameworks.
This means that the red and blue vehicle will keep their velocity and thus do not cooperate.
Fig.~\ref{fig:bottleneck_uncooperative} and Fig.~\ref{fig:merge_in_uncooperative} show the velocity 
and position graphs for the respective scenario.
The green agent $g_0$ needs to decelerate heavily in order to let the red 
vehicle pass before it can accelerate again to its desired velocity.
A similar behavior with a greater deceleration is required for the merge-in scenario, with the 
green agent being required to nearly come to a full stop.

\begin{figure}
	\centering
	\includegraphics[clip, trim= 1.2cm 1cm 2.9cm 		
	1cm,width=0.95\columnwidth]{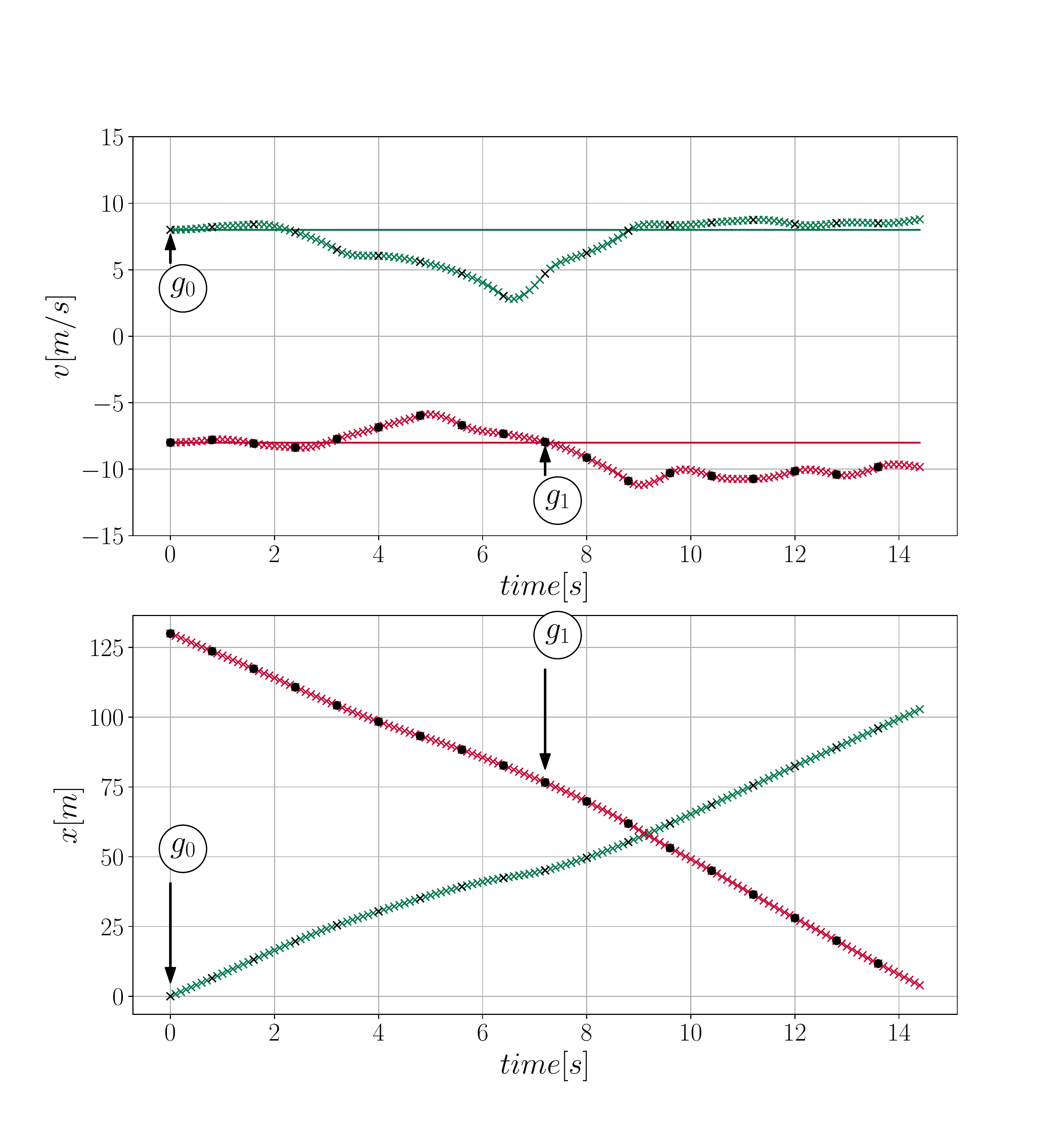}
	\caption{Bottleneck scenario with the assumption of cooperative behavior;
		agent $g_0$ and $g_1$ reduce their velocity slightly}
	\label{fig:bottleneck_cooperative}
\end{figure}

\begin{figure}
	\centering
	\includegraphics[clip, trim= 1.4cm 1cm 2.9cm 		
	1cm,width=0.95\columnwidth]{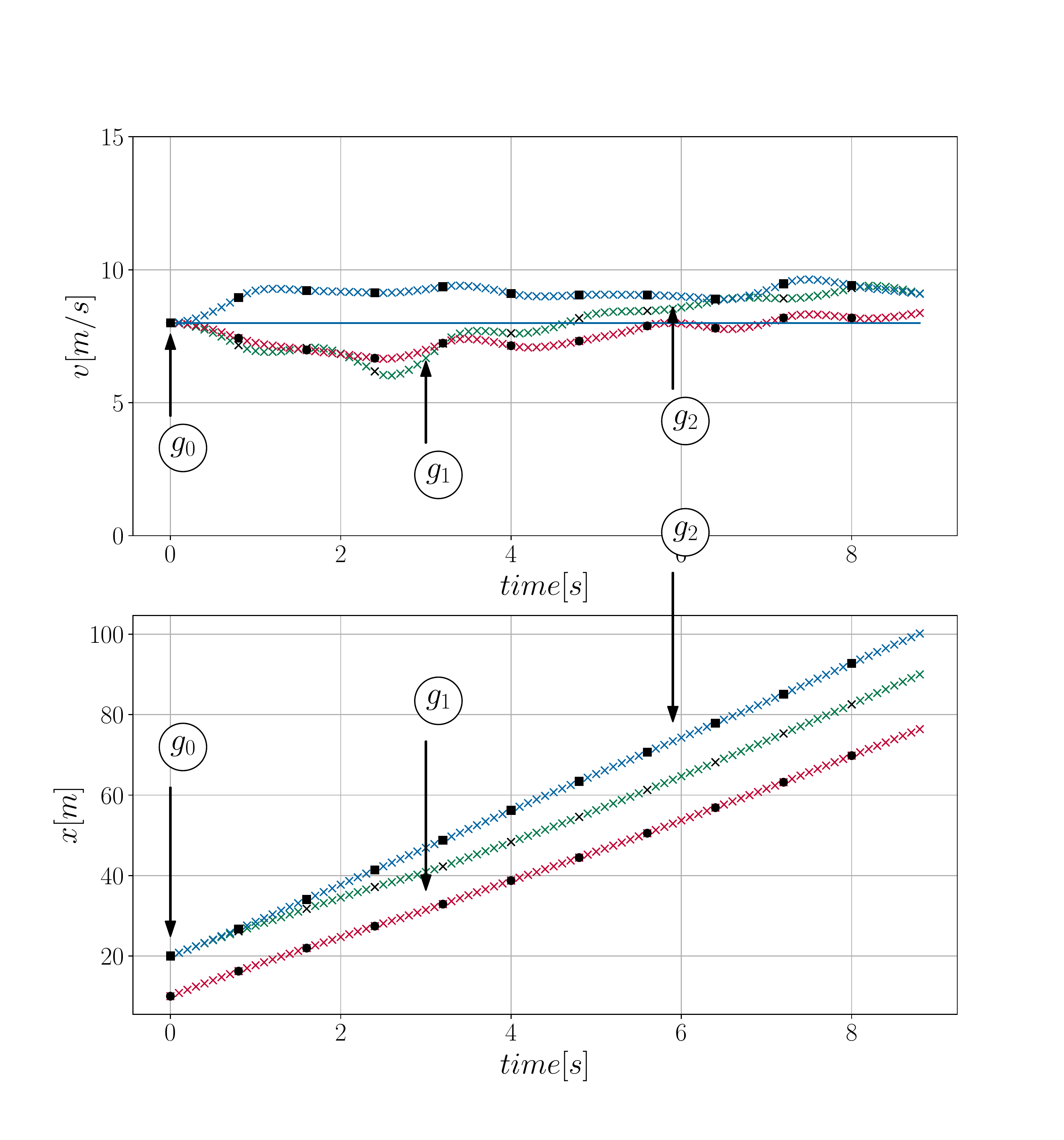}
	\caption{Merge-in scenario with the assumption of cooperative behavior;
		agent $g_2$ accelerates slightly while agent $g_1$ decelerates slightly to open a gap so that agent $g_0$ can
		merge-in with little deceleration}
	\label{fig:merge_in_cooperative}
\end{figure}

If the agent assumes that the other agents do not follow the naive constant velocity assumption, 
but rather take the perception and decision making by all traffic participants into consideration, 
a more efficient cooperative driving style can be observed.
The respective velocity and position graphs are depicted in Fig.~\ref{fig:bottleneck_cooperative} 
and Fig.~\ref{fig:merge_in_cooperative}.
During the bottleneck scenario, both vehicles decelerate only slightly and pass each other at the 
road constriction.
During the merge-in, the blue vehicle accelerates slightly while the red vehicle decelerates to 
allow the green vehicle to merge in.

It was shown, that the constant velocity assumption can lead to suboptimal solutions, that can
be avoided if one takes the interdependence of actions into account, reaching superior solutions with regards to efficiency.


\section{Conclusions}
This paper proposes a method to plan decentralized cooperative continuous trajectories for multiple 
agents to increase the efficiency of automated driving especially in urban scenarios and tight spaces.
Due to continuous action spaces arbitrary trajectory combinations can be planned, distinguishing 
this method from other cooperative planning approaches.
In order to handle the combinatorial explosion introduced by the continuous action space, we 
enhanced the MCTS-based algorithm by guided search, semantic action groups and similarity updates.
While these enhancements show promising results with regard to the exploration of the search space, 
current research focuses on tuning the methods, proving scenario independent applicability and conduct in depth quantitative analysis.

\section{Acknowledgements}
We wish to thank the German Research Foundation (DFG) for funding the project Cooperatively 
Interacting Automobiles (CoInCar)
within which the research leading to this contribution was conducted. 
The information as well as views presented in this publication are solely the ones expressed by the 
authors.

\vspace{5 cm}






\bibliographystyle{IEEEtran}
\bibliography{IEEEabrv,04_mendeley-export/library}

\end{document}